\documentclass[conference,a4paper]{IEEEtran}
\IEEEoverridecommandlockouts

\usepackage[hidelinks]{hyperref}
\usepackage[cmex10]{amsmath}
\usepackage{amssymb,amsfonts}
\usepackage{dblfloatfix}

\usepackage[ruled,vlined]{algorithm2e}
\usepackage{graphicx}
\graphicspath{{Figures/PDF/}{Figures/PNG/}}

\usepackage{booktabs}
\usepackage{siunitx}
\usepackage[numbers,compress]{natbib}
\usepackage{texnames}
\usepackage{bm,bbm}
\usepackage{orcidlink}

\usepackage{xcolor}
\usepackage{multirow}
\usepackage{amssymb}

\begin{document}

\title{A MAMBA-BASED  MULTIMODAL NETWORK FOR MULTISCALE BLAST-INDUCED RAPID STRUCTURAL DAMAGE ASSESSMENT}

\author{
    \IEEEauthorblockN{Wanli Ma, Sivasakthy Selvakumaran}
    \IEEEauthorblockA{\textit{Department of Engineering}\\ \textit{University of Cambridge}\\
    Cambridge CB2 1TN, UK\\
    \{wm369, ss683\}@cam.ac.uk}
    \and
    \IEEEauthorblockN{Dain G. Farrimond, Adam A. Dennis, Samuel E. Rigby}
    \IEEEauthorblockA{\textit{School of Mechanical, Aerospace and Civil Engineering}\\ \textit{University of Sheffield}\\
    Sheffield S10 2TN, UK\\
    \{d.farrimond, a.a.dennis, sam.rigby\}@sheffield.ac.uk}
}


\maketitle
\begin{abstract}
	Accurate and rapid structural damage assessment (SDA) is crucial for post-disaster management, helping responders prioritise resources, plan rescues, and support recovery. Traditional field inspections, though precise, are limited by accessibility, safety risks, and time constraints, especially after large explosions. Machine learning with remote sensing has emerged as a scalable solution for rapid SDA, with Mamba-based networks achieving state-of-the-art performance. However, these methods often require extensive training and large datasets, limiting real-world applicability. Moreover, they fail to incorporate key physical characteristics of blast loading for SDA. To overcome these challenges, we propose a Mamba-based multimodal network for rapid SDA that integrates multi-scale blast-loading information with optical remote sensing images. Evaluated on the 2020 Beirut explosion, our method significantly improves performance over state-of-the-art approaches. Code is available at: https://github.com/IMPACTSquad/Blast-Mamba
\end{abstract}


\section{Introduction}
Explosion-induced disasters, whether accidental or deliberate, pose serious risks to human safety and the built environment \cite{elliott1992protection, luccioni2004analysis}. Structural damage assessment (SDA) evaluates the extent, severity, and types of building damage, providing critical information for informed rescue operations \cite{berke1993recovery}. Traditionally conducted via field surveys, SDA can be highly precise, as in Zhou et al.’s \cite{zhou2022deep} rapid assessment of RC columns under blast loading. However, these methods are limited by accessibility, safety risks, and time requirements, making them inadequate for rapid and large-scale disaster response.

Advances in remote sensing imagery (RSI) have enabled rapid, large-scale SDA \cite{dong2013comprehensive, zheng2021building}, especially valuable when ground communications are disrupted after disasters. Coupled with deep learning, remote sensing supports automated detection, classification, and quantification of structural damage with high precision, as shown in earthquakes \cite{balaji2024unsupervised}, floods \cite{braik2024automated}, wildfires \cite{kang2025residential}, and hurricanes \cite{cheng2021deep}. However, supervised deep learning requires costly, time-intensive annotation and struggles to generalise across regions and sensors, limiting its effectiveness in real-world SDA. Furthermore, the integration of RSI and blast loading as critical physical information remains largely underexplored in post-disaster SDA.

To tackle these challenges, we propose a rapid multimodal SDA pipeline with pre-training and fine-tuning stages. Pre-training uses a global SDA dataset covering diverse disasters to build a foundation model. Fine-tuning then adapts this model to the target area using limited local data, while integrating blast-loading information with optical RSI to boost performance. To our knowledge, this is the first study to combine blast loading with optical RSI for rapid, large-scale SDA.
\section{Theoretical Preliminaries}
\label{sec:Theoretical_Preliminaries}

\begin{figure*}[!t]
    \centering
    \includegraphics[width=0.8\linewidth]{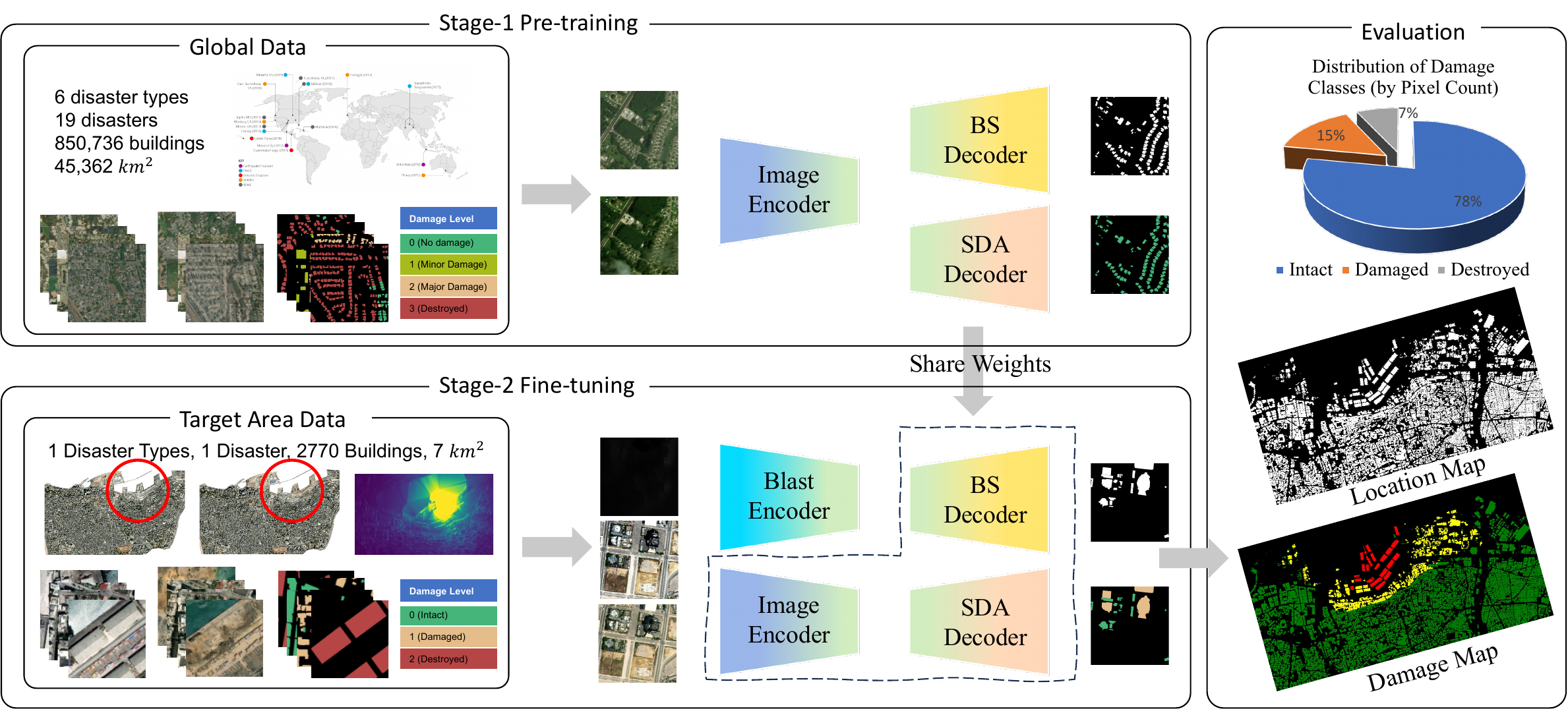}
    \vspace{-7pt}
    \caption{The workflow of the proposed rapid structural damage assessment method. BS encoder denotes the building segmentation decoder, and SDA encoder denotes the structural damage assessment decoder.}
    \vspace{-10pt}
    \label{fig:overall_framework}
\end{figure*}

\subsection{Visual State Space Model}


State space models (SSMs), exemplified by structured state space sequence models (S4) \cite{gu2021efficiently}, are linear time-invariant (LTI) systems that transform an input function $x(t) \in \mathbb{R}$ into an output $y(t) \in \mathbb{R}$ through an implicit latent state $h(t) \in \mathbb{R}^N$. The continuous system is expressed through linear ordinary differential equations (ODEs) as follows: 
$h^{\prime}(t)=\mathbf{A} h(t)+\mathbf{B} x(t)$, $
y(t)=\mathbf{C} h(t)
$, where $A \in \mathbb{R}^{N \times N}$ denotes the evolution matrix, and $B \in \mathbb{R}^{N \times 1}$ and $C \in \mathbb{R}^{1 \times N}$ denote the projection matrices.

Discretisation is necessary to integrate the continuous system into digital computation for deep learning. For example, the S4 model represents its discrete counterpart, in which a timescale parameter $\Delta$ is used to transform the continuous parameters $\mathbf{A}$ and $\mathbf{B}$ to discrete parameters $\overline{\mathbf{A}}$ and $\overline{\mathbf{B}}$. Specifically, the zero-order hold (ZOH) is employed as the discretisation rule in equation (\ref{equ:ZOH}).
\begin{equation}
\left\{\begin{array}{l}
 \overline{\mathbf{A}}=\exp (\Delta \mathbf{A}) \\
 \overline{\mathbf{B}}=(\Delta \mathbf{A})^{-1}(\exp (\Delta \mathbf{A})-\mathbf{I}) \cdot \Delta \mathbf{B} .
\end{array}\right.
\label{equ:ZOH}
\end{equation}
Then, the continuous system can be discretised as follows:
\begin{equation}
\left\{\begin{array}{l}
h_t  =\overline{\mathbf{A}} h_{t-1}+\overline{\mathbf{B}} x_t, \\
y_t  =\mathbf{C} h_t .
\end{array}\right.
\label{Equ:discretised_system}
\end{equation}
Subsequently, the computation of the recurrent model (\ref{Equ:discretised_system}) can be performed via a global convolution:
\vspace{-10pt}
\begin{equation}
\begin{aligned}
\overline{\mathbf{K}} & =\left(\mathbf{C B}, \mathbf{C} \overline{\mathbf{A B}}, \ldots, \mathbf{C} \overline{\mathbf{A}}^{\mathrm{L}-1} \overline{\mathbf{B}}\right), \\
\mathbf{y} & =\mathbf{x} * \overline{\mathbf{K}},
\end{aligned}
\end{equation}
where the input sequence $x$, with length $L$, is handled one timestep at a time. $K \in \mathbb{R}^{L}$ is a convolutional kernel.

Mamba \cite{gu2023mamba} improves SSMs by incorporating an input-dependent selection mechanism to filter out irrelevant data, while also simplifying the SSM architecture to achieve faster training and inference with reduced memory usage. Inspired by ViT \cite{dosovitskiy2020image} for visual task processing, VMamba \cite{liu2024vmamba} and Vision Mamba \cite{zhu2024vision} introduce a stack of Visual State-Space blocks with a 2D Selective Scan built upon the Mamba framework. Subsequently, ChangeMamba \cite{chen2024changemamba} employs VMamba for change detection in optical RSI. To date, ChangeMamba achieves state-of-the-art performance in post-disaster SDA across various global datasets \cite{chen2024changemamba, chen2025bright}.

\subsection{Blast Loading Generation}

The blast load information that is integrated into the SDA pipeline is simulated using the Computational Fluid Dynamics (CFD) software Viper::Blast. This GPU based solver operates with the theoretical framework given by \cite{Wada1997} and \cite{Rose2001} and has been used and validated extensively in studies focused on urban blast events (e.g. \cite{Dennis2024, Marks2021}). 

In this work, the real-world explosion in Beirut is approximately modelled using the detonation of a 0.50 kt TNT charge with a centre 10~m above the ground based on estimates from \cite{Rigby2020}. The charge is cylindrical in shape with a length to diameter aspect ratio of 1 considering the recommendation of \cite{Ritzel2023} who identified that the fireball expansion was more `egg shaped', and therefore better represented by a cylinder as opposed to a sphere or hemisphere.



\section{Methods}
\label{sec:methods}

We propose a rapid SDA method for areas affected by explosions, illustrated in Figure \ref{fig:overall_framework}. The method integrates pre- and post-event optical RSI with blast-loading information. Due to the limited size of the target area, the SDA model is first pretrained on the large-scale xBD dataset \cite{gupta2019creating}, covering 19 disaster events across six disaster types. This broad knowledge allows the model to generalise to explosion-induced damage despite scarce local data. However, differences in contextual features and data distribution necessitate fine-tuning to adapt the model to the target area, even with few samples. Incorporating blast-loading information further guides fine-tuning for explosion-specific scenarios. Thus, our pipeline consists of two stages: pre-training, which is reusable, and fine-tuning for rapid SDA in diverse target areas.

\subsection{Network Structure}

\begin{figure}
    \centering
    \includegraphics[width=\linewidth]{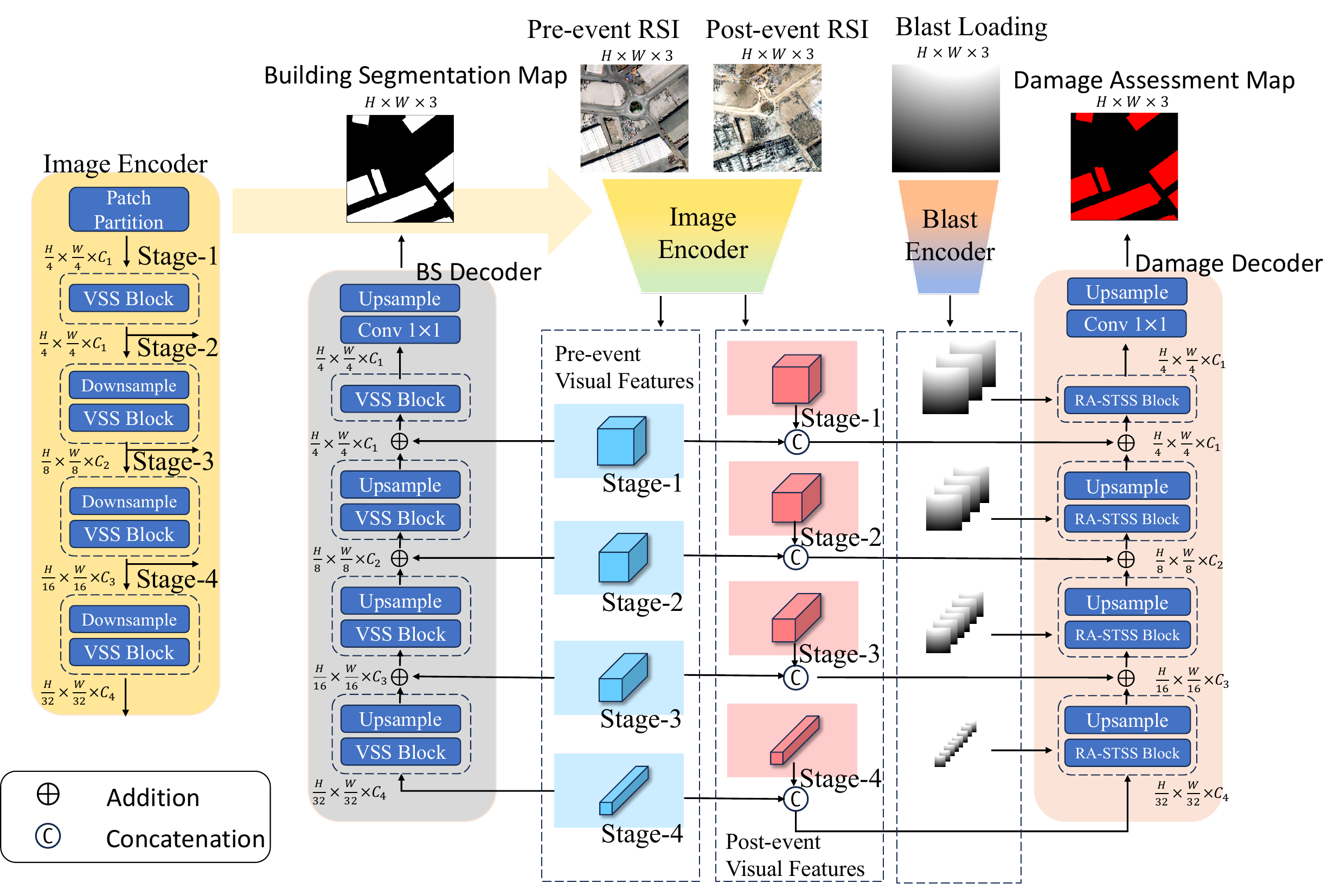}
    \vspace{-20pt}
    \caption{Structure of the fine-tuning framework. BS: building segmentation; RSI: remote sensing images.}\label{fig:network_framework}
\end{figure}

To study the impact of blast loading on structural damage, we developed a multimodal, multi-task deep learning network that jointly leverages pre- and post-event RSI along with blast-loading information. The framework, shown in Figure \ref{fig:network_framework}, builds on the VMamba and ChangeMamba models, which use an efficient scanning mechanism within the Mamba encoder for strong feature extraction. The network includes an image encoder, a blast-loading encoder, and two decoders for building segmentation (BS) and damage assessment. The BS decoder uses multi-level features from the pre-event RSI encoder via skip connections, similar to UNet, generating building masks solely from pre-event imagery. The damage assessment decoder integrates features from both pre- and post-event RSI along with blast-loading data to produce a comprehensive damage map. Our workflow, shown in Figure~\ref{fig:overall_framework}, has two steps. First, the xBD dataset (Section~\ref{subsec:Dataset}) is used to pre-train the image encoder, BS decoder, and SDA decoder. Pre-training differs from the target task: it lacks blast events, uses different damage classes, and has inconsistent image resolutions, causing a domain shift that limits performance. To address this, the model is fine-tuned on Blast-7 for Beirut 2020 explosion events, incorporating blast information via an efficient fusion strategy.

\begin{figure}
    \centering\includegraphics[width=0.8\linewidth]{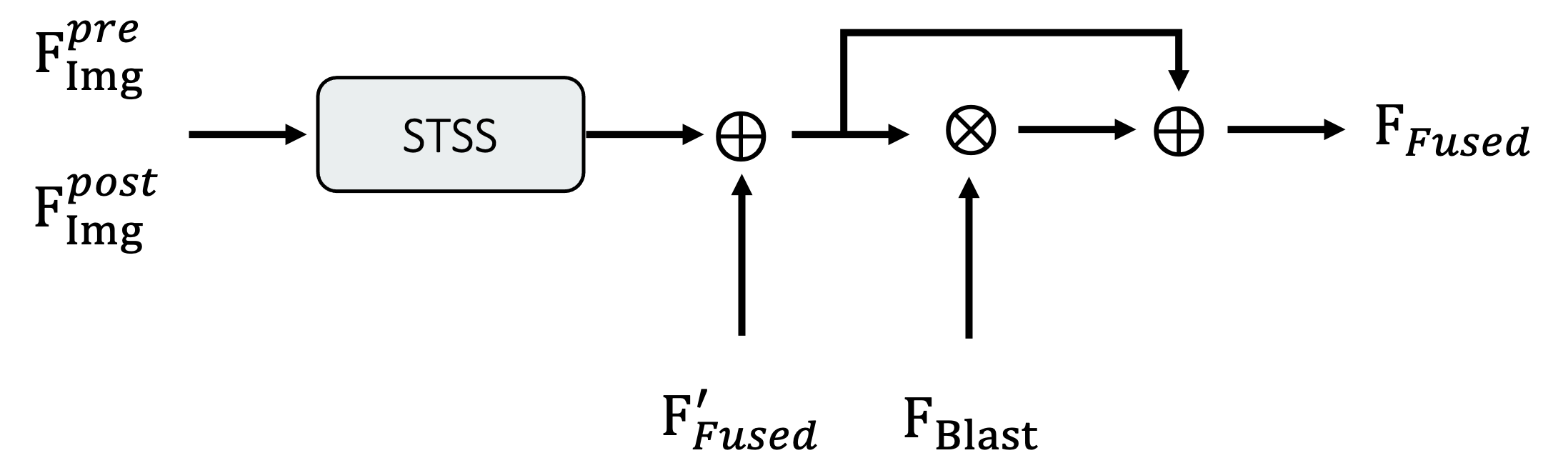}
    \vspace{-5pt}
    \caption{Residual attention–base spatiotemporal state space (RA-STSS) module}
    \vspace{-15pt}
    \label{fig:RASTSS}
\end{figure}

\textbf{Image Encoder.} Given an input image $\mathbf{I}\in\mathbb{R}^{H\times W \times 3}$,
we first apply a patch partition operation to project the image into non-overlapping patches
and map them into an embedding space:
\vspace{-10pt}
\begin{equation}
\mathbf{F}_0 = \mathrm{PatchPartition}(\mathbf{I}) 
\in \mathbb{R}^{\tfrac{H}{4}\times\tfrac{W}{4}\times C_1}.
\end{equation}
The image encoder then extracts hierarchical features through a sequence of
downsampling and visual state space (VSS) blocks:
\vspace{-10pt}
\begin{align}
\mathbf{F}_1 &= \mathrm{VSS}(\mathbf{F}_0), \\
\mathbf{F}_l &= \mathrm{VSS}(\mathrm{Down}(\mathbf{F}_{l-1})) \quad l=2,3,4.
\end{align}
The encoder is applied independently to both pre-event RSI $\mathbf{I}_{pre}$ and post-event RSI $\mathbf{I}_{post}$, yielding corresponding multi-scale features $\{\mathbf{F}_l^{pre}\}_{l=1}^4$ and $\{\mathbf{F}_l^{post}\}_{l=1}^4$.

\textbf{Mask Decoder.} To guide the network to focus on building-related regions in pre-event RSI, we adopt a mask decoder that reconstructs a binary BS map from the multi-scale encoder features:
\begin{equation}
\mathbf{M} = \mathrm{Decoder}(\{\mathbf{F}_l^{pre}\}_{l=1}^4) 
\in \mathbb{R}^{H \times W \times 3},
\end{equation}
where the $\mathrm{Decoder}$ consists of a sequence of VSS blocks followed by upsampling, as illustrated in Figure \ref{fig:network_framework}.

\textbf{Blast Encoder.} In order to incorporate the physical effect of blast loading on building damage, we introduce a blast encoder that processes the blast loading map 
$\mathbf{B} \in \mathbb{R}^{H \times W \times 3}$.
First, the blast map is interpolated to match the feature resolution $(H', W')$:
\vspace{-4pt}
\begin{equation}
\tilde{\mathbf{B}} = \mathrm{Interpolate}(\mathbf{B}, H', W'),
\vspace{-5pt}
\end{equation}
followed by a $1\times 1$ convolution to project it into the feature space and align channel dimensions of image features:
\begin{equation}
\mathbf{F^{blast}} = \mathrm{Conv}_{1\times 1}(\tilde{\mathbf{B}}).
\vspace{-5pt}
\end{equation}
Since the blast loading information is integrated into the hierarchical features of optical RSI, this process is applied at each level, producing multi-scale blast features $\{\mathbf{F}_l^{blast}\}_{l=1}^4$.

\textbf{Damage Decoder.} The damage decoder fuses multimodal inputs, pre- and post-event optical RSI and blast-loading data, for SDA. Inspired by the Spatiotemporal State Space (STSS) model \cite{chen2024changemamba}, an improved variant of VSS, we introduce a Residual Attention–based STSS (RA-STSS) block to enhance multimodal feature fusion. As shown in Figure \ref{fig:RASTSS}, pre- and post-event RSI features are concatenated, fused by STSS, combined with the previous stage’s output, and further integrated with blast-loading features via residual attention. The fused features are then decoded to generate a high-resolution damage map. Specifically:
\begin{equation}
\mathbf{U}_l = \mathrm{STSS}(cat(\mathbf{F}_l^{pre}, \mathbf{F}_l^{post}))
\quad l=1,\dots,4
\end{equation}
\begin{equation}
\mathbf{D}_l = \mathbf{U}_l \oplus \mathrm{Up}(\mathbf{U}_{l-1}) * (1 + \mathbf{F}_l^{blast})\quad l=1,\dots,4,
\end{equation}
where $\oplus$ denotes element-wise addition and we define $\mathbf{U}_{0} = 0$. Finally, a pixel-wise Argmax layer produces the structural damage map:
\vspace{-5pt}
\begin{equation}
\mathbf{Y} = \mathrm{Argmax}(\mathbf{D}_4) 
\in \mathbb{R}^{H \times W \times T},
\vspace{-5pt}
\end{equation}
where $T$ denotes the number of damage levels.

\textbf{Loss Function.} We adopt a multi-task loss to jointly supervise BS and
damage assessment. Both $\mathcal{L}_{mask}$ and $\mathcal{L}_{damage}$ are defined as
pixel-wise cross-entropy losses. The final loss is defined as $\mathcal{L} =   \mathcal{L}_{mask} +  \mathcal{L}_{damage}$.

\begin{table*}[]
    \caption{Performance comparison between the proposed method and state-of-the-art methods on the Blast-7 dataset. The symbols $\mathcal{C}$, $\mathcal{T}$, and $\mathcal{M}$ represent convolutional neural network (CNN)–based methods, Transformer-based methods, and Mamba-based methods, respectively. }
    \label{tab:placeholder}
    \centering
    \begin{tabular}{c l c c c c c c c}
\hline
\textbf{Type} & \textbf{Method} & $F_1^{\text{loc}}$ & $F_1^{\text{clf}}$ & $F_1^{\text{overall}}$ & \multicolumn{3}{c}{\textbf{Damage $F_1$ per class}} \\
\cline{6-8}
 & & & & & Intact & Damaged & Destroyed \\
\hline
\multirow{4}{*}{$\mathcal{C}$} &    DeepLabV3+ \cite{chen2018encoder}  & 81.43 & 70.61 & 73.85 & 77.66 & 53.78 & 90.76  \\
  &  UNet \cite{ronneberger2015u}  & 82.32 & 51.79 & 60.95 & 73.82 & 30.75 & 84.28  \\
  &  SiamAttnUNet \cite{adriano2021learning} &   82.05 & 60.56 & 67.01 & 71.79 & 41.79 & 85.63   \\ 
  &  SiamCRNN \cite{chen2019change}  & 85.66 & 71.74 & 75.91 & 84.90 & 51.03 & 95.73     \\ \hline
  \multirow{1}{*}{$\mathcal{T}$} & DamFormer \cite{chen2022dual}  &  85.14 & 79.54 & 81.22 & 87.03 & 63.18 & 96.16 \\
        \hline
     \multirow{2}{*}{$\mathcal{M}$} & Mamba-BDA-Small \cite{chen2024changemamba}  & 87.25 & 78.24 & 80.94 & 90.83 & 58.76 & 96.90  \\
      &
      Ours & \textbf{88.98} & \textbf{88.30} & \textbf{88.50} & \textbf{93.54} & \textbf{77.96} & \textbf{95.64} \\ 
      \hline
    \end{tabular}
\end{table*}

\section{Experiments and Results}
\label{sec:Experiments}

\subsection{Data and Experimental Setup}
\label{subsec:Dataset}
We constructed the Blast-7 dataset by combining the BRIGHT \cite{chen2025bright} dataset with blast-loading data from the 2020 Beirut explosion. To evaluate the potential of using only optical imagery, BRIGHT’s post-event images were replaced with high-resolution optical data from the Maxar Open Data Programme. Designed for rapid SDA with minimal training samples, the dataset focuses on 50 images (512 × 512 pixels) covering 7 km² around the blast centre in Beirut. Images were split into training, validation, and test sets in a 3:1:1 ratio.

The xBD \cite{gupta2019creating} is the largest building damage assessment dataset to date, derived from optical remote sensing imagery of 19 disaster events, including earthquakes, floods, volcanic eruptions, wildfires, and windstorms. 
It comprises 11,034 pairs of 1024 × 1024 remote sensing images containing 850,736 buildings, covering a total area of 45,361.79 $km^2$.

The experiments are conducted using PyTorch. The network is optimised using the AdamW optimiser, with learning rates of $1e^{-4}$ for pre-training and $1e^{-5}$ for fine-tuning. Model performance is evaluated following the protocol in \cite{chen2024changemamba, gupta2019creating} for BS and SDA, using the $F_1$ score. Specifically, $F_1^{\text{loc}}$ evaluates building localization, $F_1^{\text{clf}}$ assesses structural damage, and $F_1^{\text{overall}}$ accounts for both. Class-wise $F_1$ scores for each damage level are also presented in Section \ref{subsec:experiments}.

\subsection{Experimental Analysis}
\label{subsec:experiments}

\begin{figure}
    \centering
    \includegraphics[width=\linewidth]{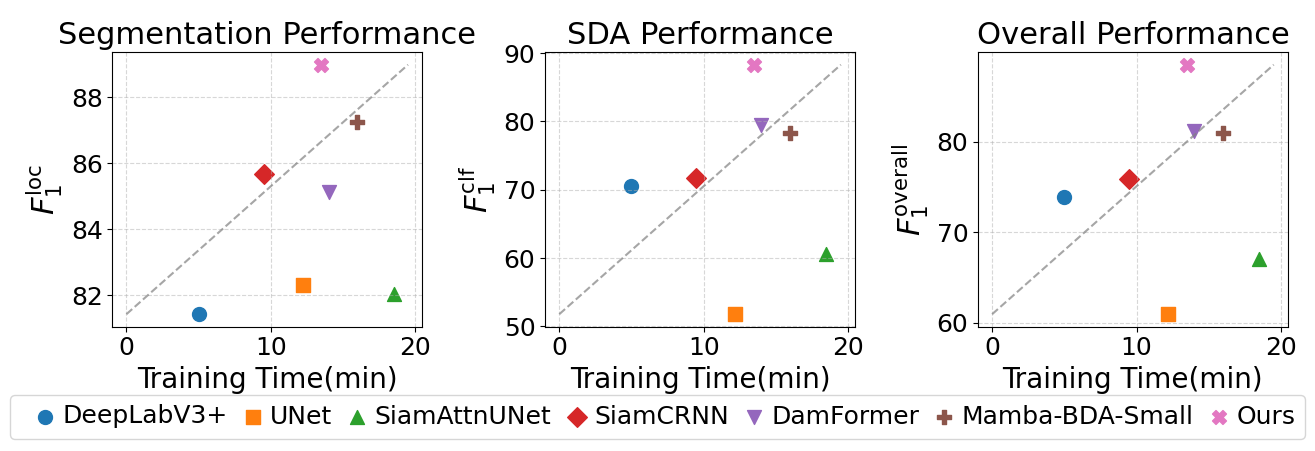}
    \vspace{-15pt}
    \caption{Comparison of training time versus performance ($F^{loc}{1}$, $F^{clf}{1}$, $F^{overall}_{1}$) of the proposed and existing methods for Blast-7 structural damage assessment.}
    \vspace{-13pt}
    \label{fig:performance_time}
\end{figure}

The proposed method was evaluated through comparison with state-of-the-art approaches, including CNN-based, Transformer-based, and Mamba-based methods. As shown in Table~\ref{tab:placeholder}, the proposed method achieves the best overall performance. Compared with CNN-based models, such as UNet and SiamCRNN, our method yields a large improvement in $F_{1}^{overall}$ (+27.55\% and +12.59\%, respectively), particularly in the ``damaged'' class where CNNs struggle (30.75\%--51.03\% compared with 77.96\%). Against the Transformer-based DamFormer, our approach achieves higher scores in both overall classification (+7.3\%) and damaged detection (+14.8\%), showing stronger ability in handling partially destroyed structures. While the Mamba-BDA baseline already improves over CNNs and Transformers, our method further enhances performance in all categories, with a notable +19.2\% gain in the ``damaged'' class, which is typically the most challenging to detect accurately. Overall, our model not only achieves the highest $F_{1}^{loc}$, $F_{1}^{cl}$, and $F_{1}^{overall}$, but also provides the most balanced accuracy across intact, damaged, and destroyed classes. In addition, Figure \ref{fig:performance_time} shows the comparison of training time versus performance
(F loc1, F clf 1, F overall
1 ) for both the proposed and existing methods in assessing Blast-7 structural damage. The proposed rapid post-disaster SDA requires only approximately 13 minutes while achieving the highest performance across all metrics.


An ablation study is conducted to evaluate the effectiveness of pretraining and the incorporation of blast loading information, and the results are reported in Table~\ref{tab:ablation}. Initial pretraining on the global xBD dataset results in weak model performance, with $F_{1}^{\text{overall}}=24.52\%$ and $F_{1}^{\text{clf}}=1.39\%$. The Step-2 fine-tuning markedly improves performance, achieving $F_{1}^{\text{clf}}=84.81\%$ and $F_{1}^{\text{overall}}=85.98\%$, with high detection rates for damaged (71.96\%) and destroyed (95.11\%) structures. Furthermore, incorporating distance information (distance from the explosion site to each point in the optical RSI) further enhances segmentation ($F_{1}^{\text{loc}}=89.09\%$) while maintaining strong overall accuracy ($F_{1}^{\text{overall}}=86.20\%$). Blast-specific fine-tuning achieves the best results ($F_{1}^{\text{clf}}=88.30\%$; $F_{1}^{\text{overall}}=88.50\%$), with superior recognition of intact (93.54\%) and destroyed (95.64\%) structures, underscoring the utility of contextual blast features for SDA.



\begin{table}[t]
\centering
\setlength{\tabcolsep}{2pt}
\renewcommand{\arraystretch}{0.85}

\caption{Ablation study on fine-tuning and the use of distance and blast information for SDA on the Blast-7 dataset. FT: fine-tuning; w/D: with distance; w/B: with blast loading.}

\label{tab:ablation}
\begin{tabular}{lcccccc}
\toprule

\textbf{Method} & $F_1^{loc}$ & $F_1^{clf}$ & $F_1^{overall}$ & \multicolumn{3}{c}{\textbf{Damage $F_1$ per class}} \\
\cline{5-7}
  & & & & Intact & Damaged & Destroyed \\
 
\midrule
Pretrain      & 78.50 & 1.39 & 24.52 & 81.09 & 0.47 & 27.70\\
FT      & 88.70 & 84.81 & 85.98 & 91.21 & 71.96 & 95.11  \\
FT w/ D      &  \textbf{89.09} & 84.96 & 86.20 & 91.55 & 71.80 & 95.58     \\
FT w/ B      & 88.98 & \textbf{88.30} & \textbf{88.50} & \textbf{93.54} & \textbf{77.96} & \textbf{95.64}\\
\bottomrule
\end{tabular}
\end{table}

\section{Conclusion}
\label{sec:conclusion}

In this paper, we present a rapid SDA solution promoting the use of blast loading information. The proposed approach involves first building a foundation model using a large global dataset, and then fine-tuning it for specific tasks in the target area. Once the foundation model is established, only the quick fine-tuning step (13 mins) is required in the target area to enable rapid SDA assessments. The results demonstrate that the proposed method outperforms the current state-of-the-art approaches, especially for the ``damaged" class, in the SDA assessment for the Beirut 2020 explosion case study. However, the proposed method still offers potential for further improvement, for instance, through fine-tuning large vision/language models to further enhance the SDA.

\section{Acknowledgment}\vspace{-0.1cm}
This work is supported by the EPSRC Project MicroBlast EP/X029018/1.

\begingroup
\footnotesize
\bibliographystyle{IEEEbib}
\bibliography{refs}
\endgroup

\end{document}